\documentclass[10pt,twocolumn,letterpaper]{article}

\usepackage{iccv}              %
\usepackage{makecell}
\usepackage{makecell} %

\definecolor{iccvblue}{rgb}{0.21,0.49,0.74}
\usepackage[pagebackref,breaklinks,colorlinks,allcolors=iccvblue]{hyperref}

\def\paperID{5230} %
\def\confName{ICCV}
\def\confYear{2025}

\usepackage{amsfonts}       %
\usepackage{nicefrac}       %
\usepackage{microtype}      %
\usepackage{mathtools}
\usepackage[numbers]{natbib}
\usepackage{lipsum}
\usepackage{array}
\usepackage{epsfig}
\usepackage{graphicx}
\usepackage{float}
\usepackage{wrapfig}
\usepackage{amsmath,amssymb,amsthm}
\usepackage{algorithm,algorithmicx,algpseudocode}
\usepackage{bm,xspace}
\usepackage{comment}
\usepackage{multirow}
\usepackage{balance}
\usepackage{url}
\usepackage{booktabs}
\usepackage{etoolbox,siunitx}
\usepackage{calc}
\usepackage{pifont,hologo}
\usepackage{color}
\usepackage{adjustbox}
\usepackage{enumitem}
\usepackage{subcaption}
\usepackage{bbding}  %
\usepackage[normalem]{ulem}  %
\PassOptionsToPackage{table}{xcolor}
\usepackage{colortbl}
\usepackage{xcolor}         %
\usepackage[table]{xcolor}

\definecolor{myblue}{HTML}{118ab2}
\definecolor{red}{HTML}{ef476f}
\definecolor{orange}{HTML}{cc7700}
\definecolor{mygray}{HTML}{efefef}
\definecolor{darkgreen}{HTML}{228B22}
\definecolor{mydarkgray}{HTML}{757575}

\renewcommand{\eqref}[1]{Eq.~\ref{#1}}

\newcolumntype{x}[1]{>{\centering\arraybackslash}p{#1}}
\newcolumntype{y}[1]{>{\raggedright\arraybackslash}p{#1}}
\newcolumntype{z}[1]{>{\raggedleft\arraybackslash}p{#1}}
\newcommand{\tablestyle}[2]{\setlength{\tabcolsep}{#1}\renewcommand{\arraystretch}{#2}\centering\footnotesize}

\DeclareMathSymbol{@}{\mathord}{letters}{"3B}

\makeatletter
\DeclareRobustCommand\onedot{\futurelet\@let@token\@onedot}
\def\@onedot{\ifx\@let@token.\else.\null\fi\xspace}

\newcommand*{\Rom}[1]{\expandafter\@slowromancap\romannumeral #1@}
\newcommand*{\rom}[1]{\expandafter\romannumeral #1}

\def\1{\bm{1}}

\let\originalleft\left
\let\originalright\right
\renewcommand{\left}{\mathopen{}\mathclose\bgroup\originalleft}
\renewcommand{\right}{\aftergroup\egroup\originalright}

\definecolor{resnet}{HTML}{264653}
\definecolor{r3m}{HTML}{264653}
\definecolor{vit}{HTML}{2A9D8F}
\definecolor{vc1}{HTML}{2A9D8F}
\definecolor{multivit}{HTML}{E9C46A}
\definecolor{multimae}{HTML}{E9C46A}
\definecolor{spunet}{HTML}{F4A261}
\definecolor{ponderv2}{HTML}{F4A261}
\definecolor{pointnet}{HTML}{E76F51}
\definecolor{cyan}{HTML}{264653}

\usepackage{pifont}%

\title{VQ-VLA: Improving Vision-Language-Action Models via Scaling Vector-Quantized Action Tokenizers}

\author{%
  Yating Wang$^{12}$ \quad  Haoyi Zhu$^{13}$ \quad Mingyu Liu$^{14}$ \quad Jiange Yang$^{15}$ \quad Hao-Shu Fang$^{6}$ \quad Tong He$^{1\textsuperscript{\dag}}$\\%\textsuperscript{\dag} Corresponding Author\\
$^1$Shanghai AI Lab \quad $^2$Tongji \quad $^3$USTC \quad $^4$ZJU \quad $^5$NJU \quad $^6$SJTU\\
\texttt{\{wangyating,liumingyu,yangjiange,hetong\}@pjlab.org.cn}\\
\texttt{\{hyizhu1108,fhaoshu\}@gmail.com}\\
$^{\textsuperscript{\dag}}$Corresponding Author
}

\begin{document}
\maketitle
\begin{abstract}

In this paper, we introduce an innovative vector quantization based action tokenizer built upon the largest-scale action trajectory dataset to date, leveraging over 100 times more data than previous approaches. This extensive dataset enables our tokenizer to capture rich spatiotemporal dynamics, resulting in a model that not only accelerates inference but also generates smoother and more coherent action outputs. Once trained, the tokenizer can be seamlessly adapted to a wide range of downstream tasks in a zero-shot manner, from short-horizon reactive behaviors to long-horizon planning. A key finding of our work is that the domain gap between synthetic and real action trajectories is marginal, allowing us to effectively utilize a vast amount of synthetic data during training without compromising real-world performance. To validate our approach, we conducted extensive experiments in both simulated environments and on real robotic platforms. The results demonstrate that as the volume of synthetic trajectory data increases, the performance of our tokenizer on downstream tasks improves significantly—most notably, achieving up to a 30\% higher success rate on two real-world tasks in long-horizon scenarios. These findings highlight the potential of our action tokenizer as a robust and scalable solution for real-time embodied intelligence systems, paving the way for more efficient and reliable robotic control in diverse application domains. \href{https://xiaoxiao0406.github.io/vqvla.github.io}{Project website}.

\end{abstract}

\section{Introduction}
\label{sec:introduction}

Tokenization plays a critical role in recent generative models, including large language models (LLMs)~\cite{gpt3,gpt4}, image and video generation models~\cite{tam2,llamagen,cogvideo}, and vision-language-action (VLA) models~\cite{brohan2023rt,kim2024openvla}. One of the key benefits of tokenization is that it compresses the input space. By reducing high-dimensional continuous data into a compact sequence of tokens, the complexity of the learning task is substantially reduced. Compared to image patches and language tokens, action sequences are inherently easier to compress because of their spatio-temporal continuity. 

Recently, several studies have started to explore action quantized tokenization for Vision-Language-Action (VLA) models~\cite{minivla,fast}, demonstrating promising potential. Effective action tokenization not only significantly enhances the downstream performance of VLA models, especially for tasks involving long-horizon planning, but also markedly improves their training and inference efficiency. In this paper, we delve deeper into the potential of action tokenization, with a specific emphasis on its scalability and accuracy.
We first find that the more precise the tokenization, the more pronounced the improvements in long-horizon action modeling. This motivates us to train the VQ tokenizer with well-scaled action trajectories to cover various tasks.
Additionally, we observe that action trajectories, unlike visual and physical modalities, exhibit minimal domain gaps between real-world and simulated environments. This characteristic enables effective scaling of action tokenizers through the extensive use of synthetic action data. Moreover, training action tokenizers is computationally lightweight compared to scaling entire VLA models. Consequently, focusing on scaling action tokenization emerges as a highly cost-effective strategy, requiring fewer computational and data resources while delivering significant performance enhancements.

Specifically, we propose a convolutional residual VQ-VAE~\cite{zeghidour2021soundstream,vqbet,minivla} framework for training action tokenizers. To effectively train the model, we propose a progressive training strategy: Initially, we train the tokenizer on real-world robotic datasets, such as OpenX-Embodiment~\cite{o2024open}, which typically contain noisy and jittery trajectories. Subsequently, we gradually integrate cleaner and smoother synthetic data from large-scale simulated robotic datasets, such as LIBERO~\cite{liu2023libero} and ManiSkill~\cite{mu2021maniskill}. This progressive approach allows the VQ model to converge toward smoother and more stable representations. Compared to previous approaches that typically rely on training with single-task datasets, our method expands the tokenizer training dataset by more than 100 times, effectively covering a broad spectrum of downstream tasks.

We conduct extensive experiments in both simulated and real-world environments. First, we evaluate our VQ-VAE action tokenizers in the LIBERO simulator, where the results demonstrate the effectiveness of the convolutional residual VQ-VAE and provide preliminary validation for the hypothesis that the VQ-VAE action tokenizer can leverage synthetic data for scaling. Additionally, real-world experiments with a Franka Research 3 robot further validate the superiority of our approach. Specifically, our findings are as follows: (1) as the amount of simulated action data increases, the VQ-VAE tokenizers exhibit linear scaling properties in improving VLA success rates; (2) the tokenizers significantly enhance inference speed and smoothness of VLA models; and (3) the tokenizers effectively reduce cumulative errors, enabling better performance in long-horizon tasks.

In summary, our contributions are as follows:
\begin{itemize}
    \item We propose a general convolutional residual VQ-VAE-based framework for action tokenizers.
    \item We demonstrate that action tokenizers can be effectively scaled by leveraging large-scale simulated action data.
    \item We prove that our action tokenizers improve the performance, inference speed, and long-horizon capabilities of VLA models.
\end{itemize}

\section{Releated Works}
\label{sec:releated works}
\textbf{Vision-Language-Action Models.}
Vision-Language-Action (VLA) models\cite{brohan2022rt,yang2023transferring,fang2023rh20t,cheang2024gr,kim2024openvla,team2024octo,zhao2025vlas,3dvla,chen2023pali,wen2024tinyvla} bridge visual-language understanding with robot control by mapping multimodal inputs to action outputs. Based on vision-language models (VLM) \cite{alayrac2022flamingo,chen2024anyvlm,chen2023pali,driess2023palm}, VLAs represent robot actions (e.g., 6DoF motion, gripper control) as discrete tokens compatible with text-based output. For example, RT-1\cite{brohan2022rt} and RT-2\cite{brohan2023rt} showed that dividing continuous actions into discrete bins allows integration with VLMs, enabling zero-shot generalization using web-scale pretraining. Recent works\cite{kim2024openvla,o2024open} further leverage this approach, demonstrating improved task generalization through large VLM backbones. Building on this foundation, recent works have explored various approaches to improve VLA, such as leveraging 3D visual inputs~\cite{3dvla}, integrating chain-of-thought reasoning~\cite{cot}, and employing parallel decoding strategies~\cite{para}. In our work, we aim to simultaneously improve the performance and execution speed of VLA.

\noindent \textbf{Action Representation.} Action representation is one of the core components in robot policy learning. Classic action representation schemes include high-level sub-tasks~\cite{ahn2022can,driess2023palm}, action primitives~\cite{pri1,vemprala2024chatgpt}, keypoints~\cite{act3d,keypoint1}, as well as low-level continuous end-effector pose~\cite{liu2023libero,o2024open} or joint state~\cite{gupta2019relay}. Recently, imitation learning-based end-to-end approaches~\cite{zhao2023learning,zhu2024point,chi2023diffusion,zhu2024spa,brohan2022rt} often utilize the latter. In order to represent complex multi-modal distributions and adapt to discrete generative models, many works further discretize continuous low-level actions, including per-dimension and per-timestep binning discretization~\cite{brohan2022rt,kim2024openvla}, VQ-VAE~\cite{vqbet,quest,minivla}, and cosine transform~\cite{fast}. Additionally, some works extract corresponding predictive signals from action-less video data to represent actions, such as pixels~\cite{cheang2024gr,stp}, trajectories~\cite{atm,tramoe}, and latent motion~\cite{ye2024latent,chen2024igor}. %

\noindent \textbf{Tokenization.} The tokenizer is one of the fundamental components in language and vision generation tasks, especially for autoregressive transformer-based generation models. For language generation tasks, mainstream methods adopt byte pair encoding (BPE)~\cite{bpe1,bpe2} to compress input text. For vision generation tasks, recent works~\cite{llamagen,tam2,taming,open,moviedreamer,diception} mainly use vector quantization~\cite{vqvae1} for tokenization to map continuous visual signals into a discrete token sequence. Similarly, in robot learning, many works~\cite{vqbet,quest,brohan2022rt,brohan2023rt,fast} model action prediction as a generative problem and tokenize the robot action modality to capture the multi-modal distribution within skills and adapt it to autoregressive generative policies. In our work, we propose a general convolutional residual VQ-VAE to tokenize robot action. Different from several releated works~\cite{vqbet,quest} that use VQ-VAE to tokenize robot actions, our VQ-VAE is trained on large-scale data and is applicable to all tasks.

\section{Methods}

\begin{figure*}[!ht]
    \centering
    \includegraphics[width=0.99\linewidth]{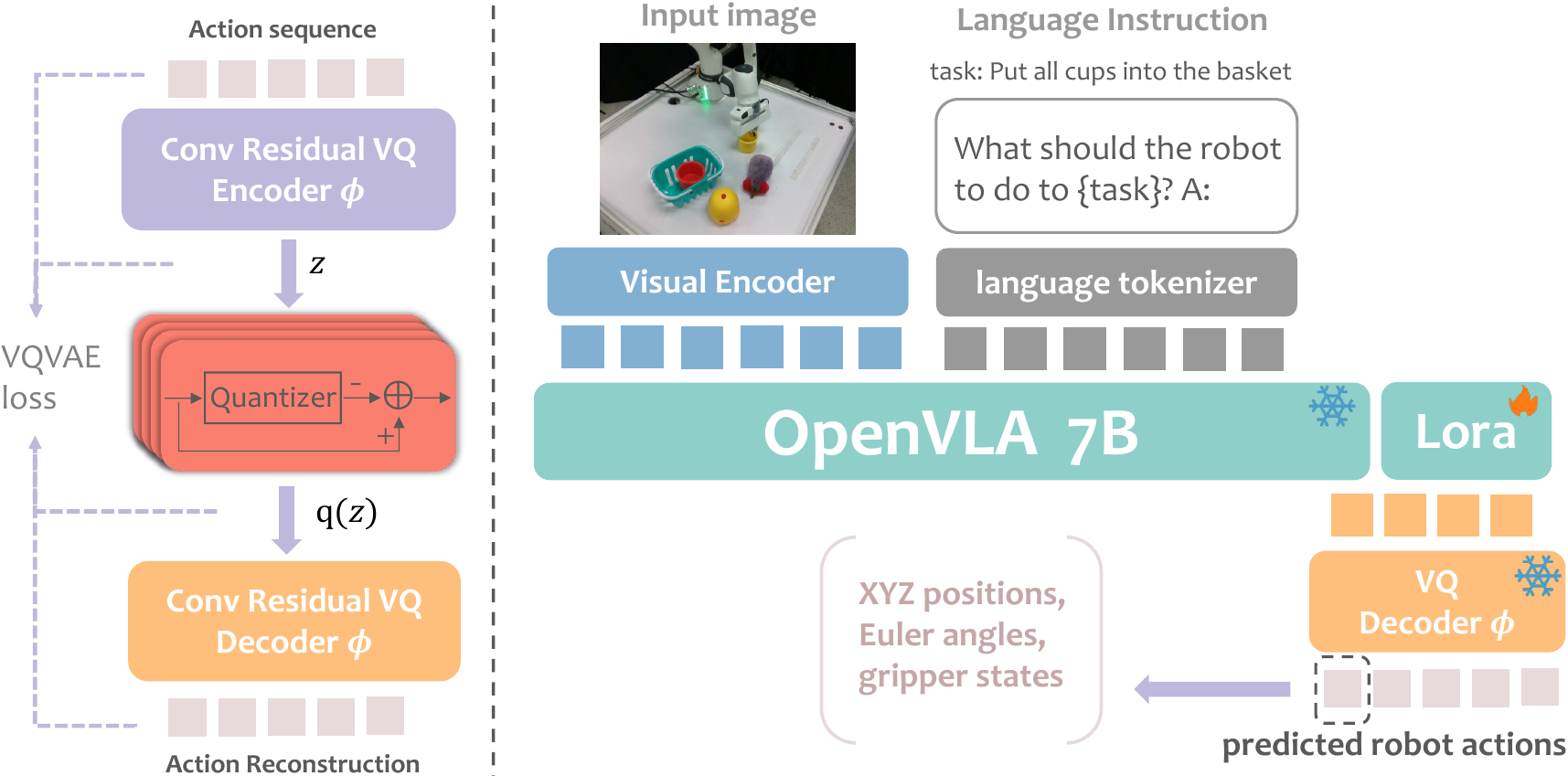}
    \caption{\textbf{The VQ-VLA pipeline}, consisting of two main stages: (1) training a general convolutional residual VQ-VAE and (2) fine-tuning OpenVLA using the LoRA approach. Specifically, a general convolutional residual VQ-VAE is first trained on the Open X-Embodiment dataset, LIEBRO, and ManiSkill datasets. The trained VQ-VAE is then frozen and serves as an action tokenizer for OpenVLA, replacing the simple binning method. In the second stage, OpenVLA is fine-tuned using the LoRA technique to optimize its performance.}
    \label{fig:enter-label}
\end{figure*}
\subsection{Preliminaries: VLA models.} We use OpenVLA\cite{kim2024openvla} as our backbone model. OpenVLA's origin formulation is to adopt a discrete tokenization strategy for robot action prediction through fine-tuning the Prismatic-7B VLM backbone. The method frames action prediction as a vision-language task, mapping input observation images and natural language instructions to discrete robot action sequences.  Specifically, continuous robot actions are discretized into 256 bins per dimension, with bin boundaries determined by the 1st and 99th percentiles of training data distributions rather than min-max ranges to mitigate outlier effects. This discretization converts N-dimensional actions into N discrete integers (0-255). To embed these into the LLM's vocabulary, OpenVLA overwrites the 256 least-used tokens in the Llama tokenizer (last 256 tokens) rather than using special tokens, as the original tokenizer only reserves 100 special tokens.

\subsection{Action Tokenizer via Residual VQ-VAE}

Based on Residual VQ-VAE~\cite{zeghidour2021soundstream,vqbet,minivla}, we design our encoder and decoder inspired by VAE in pyramidal flow matching~\cite{jin2024pyramidal}. Instead of the simple Multi-Layer Perceptron (MLP) used in Residual VQ-VAE~\cite{zeghidour2021soundstream,vqbet}, we integrate 2D temporal convolutional layers, motivated by their ability to efficiently capture local relationships and hierarchical temporal dependencies, addressing the scaling limitations of MLPs.

Given an input action sequence $\mathbf{a}_{t:t+n} \in \mathbb{R}^{n \times d}$, where $n$ is the sequence length and $d$ the action dimensionality, the encoder $\phi_\text{enc}$, composed of 2D temporal convolutional layers, transforms the sequence into a latent embedding $\mathbf{x} \in \mathbb{R}^k$, expressed as $\mathbf{x} = \phi_\text{enc}(\mathbf{a}_{t:t+n})$. To compress $\mathbf{x}$, we apply Residual Vector Quantization (RVQ)~\cite{zeghidour2021soundstream}, decomposing $\mathbf{x}$ into quantized residuals: $\mathbf{q}(\mathbf{x}) = \sum_{i=1}^{N_q} \mathbf{q}_i(\mathbf{r}_i)$, where $\mathbf{r}_1 = \mathbf{x}$, $\mathbf{r}_{i+1} = \mathbf{r}_i - \mathbf{q}_i(\mathbf{r}_i)$, and $N_q$ denotes the number of quantization stages.

The quantized embedding $\mathbf{q}(\mathbf{x})$ is passed through the decoder $\phi_\text{dec}$, which uses 2D temporal deconvolutional layers to reconstruct the sequence $\hat{\mathbf{a}}_{t:t+n}$, ensuring temporal structure preservation: $\hat{\mathbf{a}}_{t:t+n} = \phi_\text{dec}(\mathbf{q}(\mathbf{x}))$. 

To train the framework, we minimize the total loss $\mathcal{L}$, a weighted combination of reconstruction loss $\mathcal{L}_\text{rec}$, vector quantization (VQ) loss $\mathcal{L}_\text{codebook}$, and commitment loss $\mathcal{L}_\text{commit}$:
\begin{equation}
\begin{aligned}
\mathcal{L} = 
&\|\mathbf{a}_{t:t+n} - \hat{\mathbf{a}}_{t:t+n}\|_2^2 \\
&+ \lambda \big( \|\text{sg}(\mathbf{x}) - \mathbf{q}(\mathbf{x})\|_2^2 
+ \|\mathbf{x} - \text{sg}(\mathbf{q}(\mathbf{x}))\|_2^2 \big),
\end{aligned}
\end{equation}
where $\text{sg}(\cdot)$ denotes the stop-gradient operation, and $\lambda$ balances the loss components. We set $\lambda = 4$ in our experiments. This design ensures efficient, scalable, and robust encoding of structured temporal data.
\subsection{Training Residual VQ-VAE}
We train three versions of the Residual VQ-VAE to evaluate the scaling capability of our action tokenizer: (1) using only the Open X-Embodiment dataset\cite{o2024open}, (2) combining Open X-Embodiment and Libero datasets\cite{liu2023libero}, and (3) combining Open X-Embodiment, Libero, and ManiSkill\cite{mu2021maniskill} datasets. These experiments were designed to test our hypothesis that the action tokenizer can effectively scale with simulated ata.

To improve the encoder's ability to process temporal and spatial information, we introduced two types of embeddings before the action sequences are passed into the encoder:

\begin{itemize}
    \item \textbf{Time Embedding}: A sinusoidal time embedding was added to encode temporal information at varying frequencies. This embedding allows the model to capture both low-frequency and high-frequency temporal patterns in the input actions, improving its ability to represent fine-grained temporal details.
    
    \item \textbf{Action-Type Embedding}: Learnable embeddings were added for the different components of the action sequence (e.g., XYZ positions, Euler angles, and gripper states). Since the 7 dimensions of the action vector have distinct meanings, this embedding provides a strong prior for the model, helping it distinguish and process the unique roles of each dimension more effectively.
\end{itemize}

The use of these embeddings enhances the encoder's ability to process structured data, improving the quality of the latent representations and the overall performance of the tokenizer.

To train a more universal robot action tokenizer and reduce computational overhead, the model is trained using only action sequences as input, without additional conditional inputs. This design reduces complexity while maintaining the generalizability of the tokenizer. All models are trained on a single A100 GPU. For example, training on the Open X-Embodiment dataset requires just one A100 GPU and is completed in one week.

\subsection{Integrating Residual VQ-VAE as Action Tokenizer in VLA}
In this work, we replace the simple bin-based action tokenizer used in OpenVLA with a Residual VQ-VAE-based action tokenizer to improve the expressiveness and precision of action tokenization. Instead of discretizing action sequences into uniform bins, the action sequence $\mathbf{a}_{t:t+n}$ is first processed through a pre-trained and frozen Residual VQ-VAE encoder $\phi(\cdot)$, generating latent representations. These representations are then quantized into discrete codebook indices (tokens) $\{z_{q}^i\}_{i=1}^{N_q}$ corresponding to $N_q$ Residual VQ layers. The quantized tokens $z_{q}^i$ are used as the action tokens for training and prediction in the Vision-Language Model (VLM).

Unlike OpenVLA, where all token IDs are mapped to the range $[0, 255]$, the token IDs generated by different VQ layers in the Residual VQ-VAE are assigned unique, non-overlapping ranges. Specifically, tokens from the $i$-th VQ layer are offset by $(i-1) \times 256$, ensuring that:
\[
z_q^i \in [256 \times (i-1), 256 \times i - 1], \quad \forall i \in \{1, \dots, N_q\}.
\]
For example, tokens from the first layer are in the range $[0, 255]$, those from the second layer are in $[256, 511]$, and so on. This design avoids conflicts between token IDs from different layers, as tokens with the same ID in different layers represent semantically different features of the action space.

The Vision-Language Model (VLM) is trained to predict these tokens directly. The loss function is the standard next-token prediction loss, computed as the cross-entropy between the predicted token distribution $\hat{z}_{q}^i$ and the ground truth token $z_{q}^i$ produced by the frozen Residual VQ-VAE:
\[
L_{\text{VLM}} = - \sum_{i=1}^{N_q} \log P(\hat{z}_{q}^i = z_{q}^i | o_{t-h:t}),
\]
where $o_{t-h:t}$ represents the input observation sequence, and $N_q$ is the number of Residual VQ layers. 

To further enhance computational efficiency, similar to OpenVLA, the least-used tokens in the vocabulary are replaced during fine-tuning. However, this replacement process respects the layer-specific token ID ranges, ensuring that the tokenization remains consistent and interpretable across all layers.

This integration of a pre-trained and frozen Residual VQ-VAE as the action tokenizer in VLA enables a more expressive and scalable representation of actions. By leveraging hierarchical quantization with non-overlapping token ID ranges, the model achieves better action representation, avoids semantic confusion between layers, and ensures stable loss convergence during training. Furthermore, by using the VQ-VAE tokens to represent longer action sequences instead of predicting one action at a time, the model significantly reduces the number of tokens required for training and inference. This mechanism allows the model to predict more complex behavior sequences with fewer steps, leading to a substantial improvement in inference speed and computational efficiency.

\label{sec:Methods}

\section{Experiments}
In our experiments, we first validate the effectiveness and scalability of the action tokenizer in VLA models using LIBERO simulator\cite{liu2023libero}. Subsequently, real-world experiments are conducted to further verify our hypothesis. We also investigate the impact of action tokenizers on the performance, inference speed, and long-horizon capabilities of VLA models, alongside ablation studies to evaluate key design choices.

\subsection{Simulation Experiments}
\subsubsection{Experiment Setup}
We utilize the LIBERO benchmark\cite{liu2023libero} to validate and evaluate the effectiveness and scalability of the action tokenizer, using the Franka Panda robot. Specifically, the entire LIBERO task suite—including LIBERO-Spatial, LIBERO-Object, LIBERO-Goal, LIBERO-10, and LIBERO-90—is used as the entire LIBERO dataset. Among these, LIBERO-90 comprises 90 short-horizon tasks, while the other task suites each contain 10 tasks, with 50 demonstrations per task. For evaluation, testing is conducted on LIBERO-goal, LIBERO-10 and LIBERO-90.

Following the approach in OpenVLA\cite{kim2024openvla}, we filter out all "no-op" actions from the dataset. To preliminarily validate the effectiveness of using VQ-VAE as an action tokenizer in VLA models and its scalability, we train two versions of a convolutional residual VQ-VAE. The first version VQ\textsubscript{M} is trained solely on the Maniskill dataset\cite{mu2021maniskill}, and the second VQ\textsubscript{M+R} is trained on a mixture of Maniskill dataset and RLBench dataset\cite{rlbench}. Both models are trained on a single A100 GPU with a batch size of 1024, which takes about only 1 week.

We use two pre-trained VQ-VAE models as frozen action tokenizers for OpenVLA: VQ\textsubscript{M} and VQ\textsubscript{M+R} . LoRA is then applied to fine-tune OpenVLA on the LIBERO-90 dataset. Additionally, we fine-tune the original OpenVLA model on the LIBERO-90 dataset using LoRA as a baseline for comparison.For a fair comparison, all fine-tuning on the LIBERO-90 task suite is conducted for 400K gradient steps with a batch size of 4, using 4 A100-80GB GPUs and an action chunk length of K=5.

\subsubsection{Effectiveness of Conv Residual VQ-VAE}

In the early stages of the experiment, we selected two task suites, LIBERO-10 and LIBERO-GOAL, to conduct preliminary scaling validation experiments. Specifically, we used two variants of Residual VQ-VAE models: one with a simple MLP as the encoder and decoder, and the other with a larger 2D temporal convolutional network as the encoder and decoder. The results are summarized in Tab.~\ref{encoder_lierbo_result}. The findings indicate that using temporal convolutional networks as the encoder and decoder in the Residual VQ-VAE significantly outperforms the MLP-based architecture in terms of success rate. These results suggest that temporal convolutional networks can serve as an effective action tokenizer, capable of capturing temporal dependencies more effectively.

Furthermore, when we increased the training data for the VQ-VAE model, transitioning from data solely derived from individual LIBERO tasks to the entire LIBERO dataset, the success rate consistently improved. This observation provides preliminary evidence supporting the scalability of our action tokenizer design.
\begin{table}[h]
    \centering
    \tablestyle{0.25pt}{1.05}
    \resizebox{\columnwidth}{!}{%
    \begin{tabular}{l|l|c|c}
        \toprule
        \textbf{} & \textbf{Training Dataset of VQ} & \textbf{LIBERO-10 (\%)} & \textbf{LIBERO-GOAL (\%)} \\
        \midrule
        Original OpenVLA & - & 51.0 & \textbf{75.8} \\
        \midrule
        \multirow{3}{*}{MLP Residual VQ-VAE} 
        & ALL-LIBERO & 53.4 & 72.6 \\
        & LIBERO-10  & 53.2 & - \\
        & LIBERO-GOAL & - & 65.2 \\
        \midrule
        \multirow{3}{*}{Conv Residual VQ-VAE} 
        & ALL-LIBERO & \textbf{60.0} & \underline{75.2} \\
        & LIBERO-10  & \underline{54.0} & - \\
        & LIBERO-GOAL & - & 72.4 \\
        \bottomrule
    \end{tabular}%
    }
    \caption{\textbf{The evaluation results of residual VQ-VAE architectures.}The results demonstrate that the Conv Residual VQ-VAE outperforms the MLP-based version, particularly when trained on the full LIBERO dataset (ALL-LIBERO), highlighting its ability to better capture temporal dependencies and improve success rates.}
    \label{encoder_lierbo_result}
\end{table}

\subsubsection{Scaling Data Improves VQ-VAE Action Tokenizer Performance}

In the experiment, to avoid potential performance inflation from in-domain data during evaluation on LIBERO-90, we trained the VQ-VAE Action Tokenizer on completely out-of-domain ManiSkill and RLBench simulation data, rather than LIBERO data. we compare three models: the baseline, which fine-tunes the original OpenVLA; VQ\textsubscript{M}, which uses a VQ-VAE trained on the Maniskill dataset as the action tokenizer for OpenVLA; and VQ\textsubscript{M+R}, which uses a VQ-VAE trained on a mixture of the Maniskill and RLBench datasets as the action tokenizer for OpenVLA.

On LIBERO-90, VQ\textsubscript{M+R} achieved $80.98\%$, a $7.45\%$ improvement over the OpenVLA baseline ($73.53\%$), Tab.~\ref{tab:sim_result}. Furthermore, an ablation study using only ManiSkill data for training VQ\textsubscript{M} resulted in substantially lower performance, underscoring the critical role of sufficient synthetic data scale. 
\begin{table}[h!]
\centering
\tablestyle{0.25pt}{1.05}
\begin{tabular}{l|c|c|c}
\toprule
                & \makecell{\textbf{  baseline}(\%)  } 
                    & \makecell{\textbf{VQ\textsubscript{M}}(\%)} 
                    & \makecell{\textbf{VQ\textsubscript{M+R}}(\%)} \\
\midrule
\textbf{LIEBRO-90}        & 73.53  & 14.38   & \textbf{80.98} \\
\bottomrule
\end{tabular}
\caption{\textbf{Effectiveness of VQ-VAE Action Tokenizers in Scaling Simulation Data.}The results demonstrate VQ\textsubscript{M+R} reached 80.98\%, outperforming the OpenVLA baseline by 7.45\% }
\label{tab:sim_result}
\end{table}

\subsection{Real-Word Experiment}
\subsubsection{Experiment Setup}
Our robotic platform consists of a single Franka Research3 arm equipped with a third-person-view RealSense D435 camera mounted in a fixed position to capture environmental observations. The system operates at 20 Hz (moderately reduced from the native 100 Hz control frequency to balance training efficiency and motion continuity), with actions defined as absolute end-effector poses in SE(3) space (3D position + quaternion orientation). 

\begin{figure}[!ht]
    \centering
    \includegraphics[width=\linewidth]{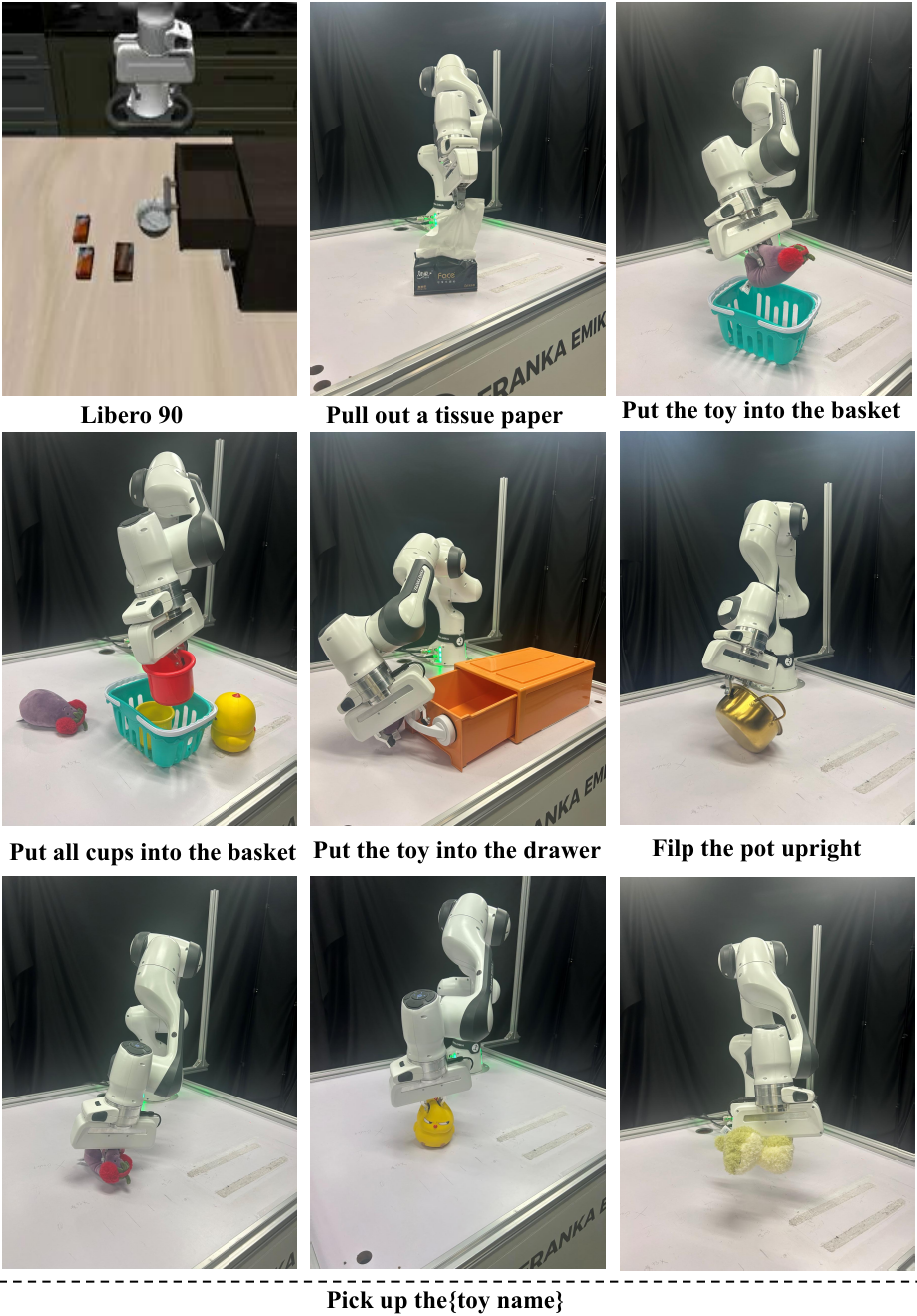}
    \caption{\textbf{All Evaluation environments}:We conduct comprehensive evaluations of VQ-VLA in both simulation and real-world settings. In simulation, evaluations are performed on the LIBERO-90 benchmark within the LIBERO dataset. And six diverse tasks are designed for real-world testing.}
    \label{fig:exp_setup}
\end{figure}
Our experimental benchmark comprises six manipulation tasks (4 short-horizon tasks, 2 long-horizon tasks) designed to evaluate the model's ability to handle varying task complexities. For each task, we collect 50 demonstrations and evaluate performance over 20 trials:

1) \textbf{Pull out a tissue paper}: The robot need to grasp and pull out a single tissue paper.

2) \textbf{Pick up the [TOY NAME]}:The robot is required to pick up the specific toy, including the toy snake, the toy eggplant, and the toy chicken, resulting in a total of three tasks (with no other distractions).

3) \textbf{Put the toy into the basket}: The robot need to pick up the toy(no other distractions) and put it into the basket.

4) \textbf{Flip the pot upright}: We set a flipped pot on the platform, the robot need to flip and upright a fallen cooking pot.

5) \textbf{Put all cups into the basket}: We place two different cups on the table and set a few other things(toys) as distractions. The robot need to sequentially put two cups into the basket, testing long-horizon task.

6) \textbf{Put the toy into the drawer}: We place a toy on the table and set a few other things as distractions. The robot need to sequentially open the drawer,pick up the specific toy and put it into the drawer, and close the drawer, testing long-horizon task.

We finetune each task separately for 100K gradient steps (batch size 4 across 4 A100-80GB GPUs) with action chunk length K=5. During deployment. 

\begin{figure*}[!ht]
    \centering
    \includegraphics[width=\linewidth]{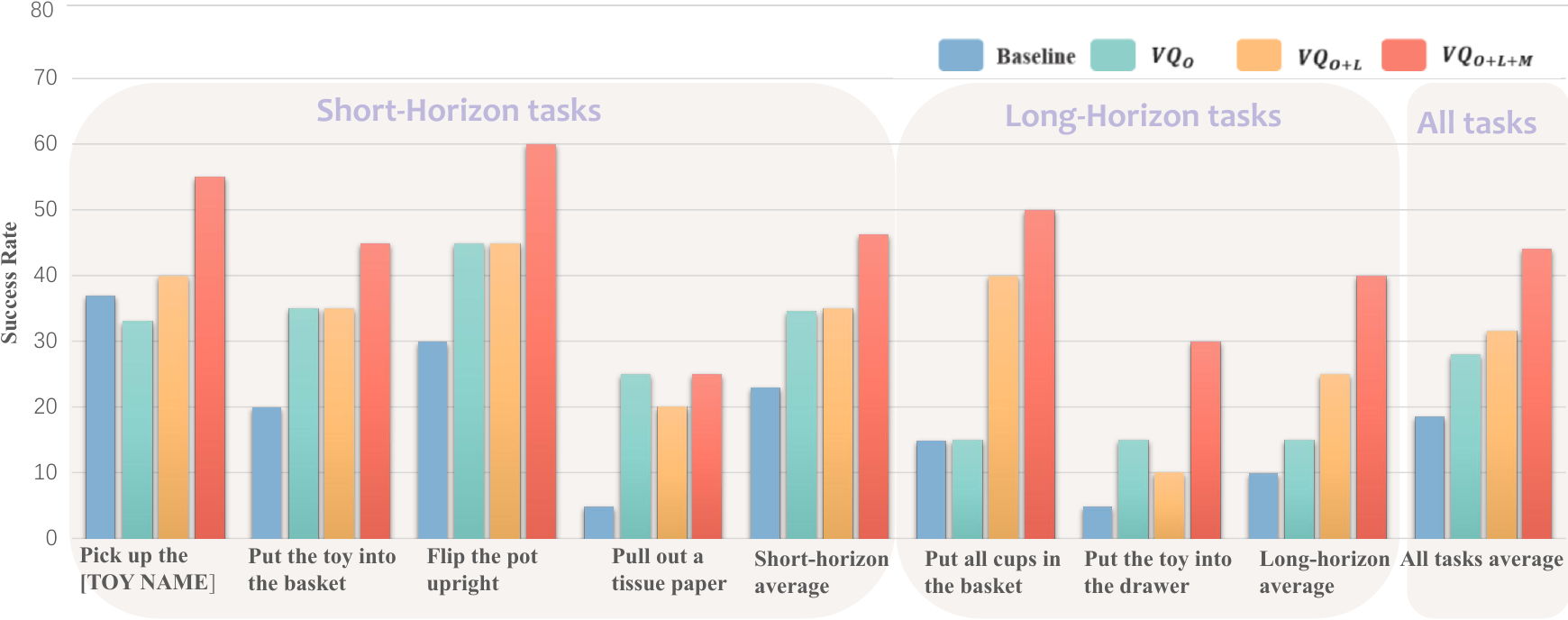}
    \caption{\textbf{Real-world experimental results:} We compare the performance of Baseline, VQ\textsubscript{O}, VQ\textsubscript{O+L}, and VQ\textsubscript{O+L+M} on both short-horizon and long-horizon tasks. In terms of the average success rate, all VQ-based models outperform the Baseline. The best-performing model, VQ\textsubscript{O+L+M}, achieves a success rate that is 23.25\% higher than the Baseline on both short-horizon and long-horizon tasks. Additionally, the results show that VQ\textsubscript{O+L+M} outperforms VQ\textsubscript{O+L}, which in turn outperforms VQ\textsubscript{O}, indicating the effectiveness of incorporating synthetic data during training without compromising real-world performance.}
    \label{fig:real_world_results}
\end{figure*}

\subsubsection{Performance on Short-Horizon Tasks}
Based on the pre-trained VQ\textsubscript{O} and VQ\textsubscript{O+L} in the simulation experiments, we further incorporated 120k synthetic trajectories from the ManiSkill dataset, co-training VQ\textsubscript{O+L+M} with Open X-Embodiment dataset and LIBERO dataset to evaluate the scaling capability of our action tokenizer. We evaluated the performance of the action tokenizer trained with different datasets on four carefully designed short-horizon tasks, focusing on multi-task learning, precision manipulation, and dynamic adjustment capabilities. The results are shown in Fig.~\ref{fig:real_world_results}. 

It can be observed that when synthetic trajectories are added for co-training, the performance of the action tokenizer improves significantly. The average success rate increased from the baseline of 23\% to 46.25\%, with a notable improvement of 30\% in the "Flip the pot upright" task. In the "Pull out a tissue paper" task, which tests the robot's performance in high-precision dynamic operations (as this task requires continuous, fine-grained grasping and pulling motions), the baseline model achieved only a 5\% success rate, failing almost entirely. In contrast, models incorporating VQ achieved success rates of 20\% or higher. 

Furthermore, comparing VQ\textsubscript{O+L} with VQ\textsubscript{O}, the average success rate on short-horizon tasks increased by only 0.5\%. This may be attributed to the limited size of the LIBERO dataset, which has minimal impact on short-horizon tasks. However, the ManiSkill dataset is 50 times larger than LIBERO, leading to a significant improvement in success rates when used for training.

\subsubsection{Performance on Long-Horizon Tasks}

VQ-VLA demonstrates outstanding performance on long-horizon tasks ("Put all cups in the basket" and "Put the toy into the drawer"), significantly outperforming baseline model in both success rate and efficiency. In scenarios where baseline models achieve success rates as low as 15\% or nearly 0, the VQ$_{O+L+M}$ model achieves significantly higher success rates of 50\% and 30\%, respectively (see Fig.\ref{fig:real_world_results}). For the "Put the toy into the drawer" task, a representative long-horizon scenario, the baseline model was only able to complete the first step of opening the drawer in 15 out of 20 trials, failing to proceed further in most cases. In contrast, the VQ$_{O+L+M}$ model successfully opened the drawer in all test cases, demonstrating its robustness and reliability in handling complex sequential tasks.

A key advantage of VQ-VLA in long-horizon tasks lies in its use of VQ-VAE as an action tokenizer, which enables the model to predict multiple actions in a single inference step. This design reduces the accumulation of errors over extended task sequences, making VQ-VLA particularly advantageous for tasks requiring long-term planning and execution. Additionally, in scenarios involving the consecutive execution of multiple subtasks, VQ-VLA not only achieves higher success rates but also significantly reduces task completion time compared to baseline models, highlighting its superior execution efficiency and ability to handle long-horizon challenges effectively.

\subsubsection{Sim\&Real Domain Gap Analysis}
To quantify the domain gap between synthetic data and real-world data, we trained a VQ-VAE model exclusively on the LIBERO dataset, referred to as VQ\textsubscript{L}, for OpenVLA. The model was tested in three real-world tasks (one long-horizon and two short-horizon tasks), and the results are shown in Tab.~\ref{tab:libero_vq}. The results indicate that the performance of VQ\textsubscript{L} is comparable to that of both VQ\textsubscript{O+L} and VQ\textsubscript{O}, suggesting that the domain gap between synthetic and real-world data is minimal. 

Although real-world data may contain noise, the inclusion of Open X-Embodiment data as a real-world dataset expands the data sources and enriches the diversity of data types, which effectively enhances the model's generalization and robustness.

\begin{table}[h!]
\centering
\resizebox{\columnwidth}{!}{ %
\begin{tabular}{l|c|c|c}
\toprule
                    & \makecell{\textbf{Put the toy}\\\textbf{into the drawer (\%)}} 
                    & \makecell{\textbf{Flip the pot}\\\textbf{upright (\%)}} 
                    & \makecell{\textbf{Put the toy}\\\textbf{into the basket (\%)}} \\
\midrule
\textbf{baseline}   & 5.0      & 30.0               &20.0                          \\
\textbf{VQ\textsubscript{O}}        & 15.0        & 45.0             & 35.0                        \\
\textbf{VQ\textsubscript{L}}     & 10.0        & 55.0     & 35.0                         \\
\textbf{VQ\textsubscript{O+L}} & 10.0       & 45.0             & 35.0                        \\
\textbf{VQ\textsubscript{O+L+M}} & \textbf{25.0} & \textbf{60.0}               & \textbf{45.0}               \\
\bottomrule
\end{tabular}}
\caption{\textbf{Performance Comparison Across Real-World Tasks:} We observe that the performance of VQ\textsubscript{L} is comparable to that of both VQ\textsubscript{O+L} and VQ\textsubscript{O}, indicating that the domain gap between synthetic and real-world data is minimal.}
\label{tab:libero_vq}
\end{table}

\subsubsection{Inference Speed Comparison}
During the real-world experiments, we measured the action execution frequency of VQ-VLA and compared it with the original OpenVLA. The results are summarized in Tab.~\ref{tab:frequency}. As shown in the table, with a compression ratio of 5 in VQ-VAE, the inference speed is nearly tripled. This significant improvement greatly facilitates real-time performance in practical applications.

\begin{table}[h!]
\centering
\tablestyle{29.5pt}{1.05}
\begin{tabular}{l|c} %
\toprule %
 & Frequency (Hz) \\ %
\midrule %
VQ-VLA & \textbf{11.84} \\ %
OpenVLA & 4.16 \\ %
\bottomrule %
\end{tabular}
\caption{\textbf{The Results of Frequencies.} We report the comparison results of our VQ-VLA and baseline OpenVLA.}
\label{tab:frequency}
\end{table}

\subsection{Ablation Studies}

In this section, we report some ablation studies to show the effectiveness of the design choices of our method.

\subsubsection{Action Chunking via VQ-VAE and Autoregressive Output}

The original OpenVLA model generates a single action per inference step based solely on the current observation, resulting in a step-by-step execution approach. To improve efficiency and handle longer action sequences, we extend OpenVLA to output a sequence of five actions in an autoregressive manner, similar to the action chunking mechanism in VQ-VLA. Specifically, the model predicts the next action based on the current observation and previously generated actions, allowing it to generate a sequence of actions in a single inference step. This autoregressive approach serves as an alternative action chunking strategy by grouping multiple actions for execution.

To evaluate the effectiveness of different action chunking strategies, we design ablation experiments comparing the autoregressive output of OpenVLA to the VQ-based action chunking method in VQ-VLA. The results, shown in Tab.~\ref{tab:action_chunk}, indicate that the use of the autoregressive approach for action chunking in the original OpenVLA leads to a significant drop in success rate compared to the baseline.
In real-world experiments, when using the autoregressive approach as a form of action chunking, the spatial magnitude of each action chunk is relatively small, resulting in slower overall execution of actions. Compared to the Action Chunking via VQ-VAE method, even with the same action chunk size, the time required to reach the same target location is significantly longer. Additionally, this approach tends to exhibit shortcut learning, where the model directly copies previous actions within the same action chunk. By analyzing the output values of each action chunk, it is observed that multiple actions within a single chunk have remarkably similar values, indicating a lack of diversity in the predicted actions. 

Overall, Action Chunking via VQ-VAE not only improves inference speed but also enhances the performance of VLA by generating more effective and diverse action sequences. This makes it a more suitable approach for addressing real-world long-horizon tasks.

\begin{table}[h!]
\centering
\resizebox{\columnwidth}{!}{ %
\begin{tabular}{l|c|c|c}
\toprule
    \textbf{Action Chunking}                & \makecell{\textbf{LIBERO-90 (\%)}} 
                    & \makecell{\textbf{Flip the pot}\\\textbf{upright (\%)} } 
                    & \makecell{\textbf{Put the toy}\\\textbf{into the basket (\%)}} \\
\midrule
\textbf{baseline}   & \underline{74.76}       & \underline{30.0}               & \underline{20.0}                          \\
\textbf{Autoregressive Output}        & 66.53        & 10.0              & 0.0                       \\
\textbf{VQ-based (VQ\textsubscript{O+L+M})}   & \textbf{86.61}       & \textbf{60.0}               & \textbf{45.0}                          \\
\bottomrule
\end{tabular}}
\caption{\textbf{Comparison of action chunking methods:} We evaluate the performance on three tasks (one simulator task and two real-world tasks). The results show that the Autoregressive Output used in OpenVLA performs poorly as an action chunking method, with a significant gap compared to the VQ-based approach. This indicates that the VQ-based method is more effective.}
\label{tab:action_chunk}
\end{table}

\subsubsection{Embedding Integration Effectiveness}
To evaluate the impact of embeddings, we conducted an ablation study comparing the model's performance with and without embeddings. Specifically, the baseline model processes raw action sequences directly, while the enhanced model incorporates both time embedding and action-type embedding. 

As shown in Tab.~\ref{tab:embedding_ablation}, the model with embeddings significantly outperforms the baseline in terms of success rate. This demonstrates that the integration of embeddings improves the encoder’s ability to represent structured action sequences, leading to better overall performance.

\begin{table}[h!]
\centering

\resizebox{\columnwidth}{!}{ %
\tablestyle{0.25pt}{1.05}
\begin{tabular}{l|c|c|c}
\toprule
                & \makecell{\textbf{LIBERO-90 (\%)}} 
                    & \makecell{\textbf{Flip the pot}\\\textbf{upright (\%)}} 
                    & \makecell{\textbf{Put the toy}\\\textbf{into the basket (\%)}} \\
\midrule
\textbf{VQ\textsubscript{O+L} (w.o. Embeddings)}        & 85.17        & 40.0              & 35.0                       \\
\textbf{VQ\textsubscript{O+L} (w Embeddings) }   & \textbf{86.16}       & \textbf{45.0}               & \textbf{35.0}                          \\

\bottomrule
\end{tabular}}
\caption{\textbf{Embedding integration improves performance.} The table compares models with and without embeddings across three tasks, showing that embeddings enhance success rates, especially for "Flip the pot upright."}
\label{tab:embedding_ablation}
\end{table}

\section{Limitations and Future Works}
In our work, extensive experiments demonstrate that our proposed action tokenizers can improve VLA performance and inference speed while achieving scalability on simulated data. However, there still remain some limitations and opportunities for future work. First, our action tokenizers can be further extended to larger-scale simulated datasets, such as RLbench~\cite{rlbench} based on CoppeliaSim. Second, our work improves inference speed by decoding multi-step action sequences, which can be further combined with techniques such as distillation~\cite{ddd} and quantization~\cite{qqq} of VLMs in the future. Fially, the architecture design of our action tokenizers can be further improved, for example, by incorporating the frequency of action data as a extra condition. In summary, we hope that our work can serve as a strong baseline and inspire future research.

\section{Conclusions}

In this paper, we propose a general convolutional residual VQ-VAE framework for action tokenizers that can seamlessly integrate with state-of-the-art VLA models. Our VQ-VAE is trained on 100 times more data than previous methods and can be directly transferred to downstream both real-world and simulated robotic manipulation tasks. Extensive experiments conducted in both simulated and real-world environments validate that the proposed convolutional residual VQ-VAE framework not only enhances the performance of VLA policies but also accelerates inference. Finally, our VQ-VAE also demonstrates the ability to scale effectively with large-scale simulated data.

\section*{Acknowledgments}
This work is supported by the National Key R\&D Program of China (NO.2022ZD0160102) and Shanghai Artificial Intelligence Laboratory.

{
    \small
    \bibliographystyle{ieeenat_fullname}
    \bibliography{main}
}

\end{document}